%% file: GoDiff.tex
\documentclass[letterpaper]{article} 
\usepackage{aaai25}  
\usepackage{times}  
\usepackage{helvet}  
\usepackage{courier}  
\usepackage[hyphens]{url}  
\usepackage{graphicx} 
\usepackage{natbib}  
\usepackage{caption} 
\usepackage{algorithm}
\usepackage{algorithmic}

\usepackage{newfloat}
\usepackage{listings}
\DeclareCaptionStyle{ruled}{labelfont=normalfont,labelsep=colon,strut=off} 
\floatstyle{ruled}
\newfloat{listing}{tb}{lst}{}
\floatname{listing}{Listing}
\usepackage[flushleft]{threeparttable}

\usepackage{array}
\newcolumntype{I}{!{\vrule width 3pt}}
\newlength\savedwidth

\newlength\savewidth
\newcommand\shline{\noalign{\global\savewidth\arrayrulewidth
\global\arrayrulewidth 1.25pt}%
\hline
\noalign{\global\arrayrulewidth\savewidth}}

\usepackage{multirow}
\usepackage{makecell}
\usepackage{adjustbox}
\usepackage{booktabs}

\usepackage{graphicx}

\usepackage{colortbl}
\usepackage{xcolor}
\usepackage{amsmath}
\usepackage{amssymb}
\usepackage{pifont}
\usepackage{xcolor}

\title{Object Style Diffusion for Generalized Object Detection in Urban Scene}
\author {
    Hao Li\textsuperscript{\rm 1},
    Xiangyuan Yang\textsuperscript{\rm 2},
    Mengzhu Wang\textsuperscript{\rm 1},
    Long Lan\textsuperscript{\rm 1}, 
    Ke Liang\textsuperscript{\rm 1}, \\
    Xinwang Liu\textsuperscript{\rm 1}\thanks{Xinwang Liu is the corresponding author.},
    Kenli Li\textsuperscript{\rm 3}
}
\affiliations {
    \textsuperscript{\rm 1}National University of Defense Technology, China\\
    \textsuperscript{\rm 2}Xi'an Jiaotong University, China\\
    \textsuperscript{\rm 3}Hunan University, China\\
    (lihao96@nudt.edu.cn, dreamkily@gmail.com; long.lan@nudt.edu.cn; liangke200694@126.com;  xinwangliu@nudt.edu.cn).
}

\usepackage{bibentry}

\begin{document}

\maketitle

\begin{abstract}
Object detection is a critical task in computer vision, with applications in various domains such as autonomous driving and urban scene monitoring. However, deep learning-based approaches often demand large volumes of annotated data, which are costly and difficult to acquire, particularly in complex and unpredictable real-world environments. This dependency significantly hampers the generalization capability of existing object detection techniques. To address this issue, we introduce a novel single-domain object detection generalization method, named GoDiff, which leverages a pre-trained model to enhance generalization in unseen domains. Central to our approach is the Pseudo Target Data Generation (PTDG) module, which employs a latent diffusion model to generate pseudo-target domain data that preserves source domain characteristics while introducing stylistic variations. By integrating this pseudo data with source domain data, we diversify the training dataset. Furthermore, we introduce a cross-style instance normalization technique to blend style features from different domains generated by the PTDG module, thereby increasing the detector's robustness. Experimental results demonstrate that our method not only enhances the generalization ability of existing detectors but also functions as a plug-and-play enhancement for other single-domain generalization methods, achieving state-of-the-art performance in autonomous driving scenarios.

\end{abstract}

\input{sections/Introduction}

\input{sections/Related_Work}

\input{sections/Method}

\input{sections/Expriments}

\input{sections/Conclusion}


\bibliography{ref}

\clearpage
\input{sections/Appendix}

\end{document}


\maketitle


%


%% file: sections/Introduction.tex
\section{Introduction}

Object detection technologies, particularly through deep learning, are integral to various computer vision applications, including autonomous driving and urban surveillance \cite{obreview}. These technologies empower computer vision systems to accurately identify and localize objects within images and videos, thereby supporting subsequent tasks like decision-making. However, the diversity and unpredictability of real-world scenarios pose significant challenges to the generalization capability of object detection models~\cite{adaptation2020}. Moreover, deep learning-based object detection models require extensive annotated data, which are costly to produce and often insufficient to encompass all potential scenario variations~\cite{dadapt}. Consequently, improving the generalization ability of object detection models in unseen scenarios becomes a practical yet challenging task in this field.

To address this challenge, researchers have introduced Single-Domain Generalization (S-DG), a novel task aimed at generalizing models across multiple unknown target domains using only a single source domain~\cite{sdg}. In this context, data augmentation is a key strategy for enhancing model generalization by diversifying training data, thereby simulating unseen scenarios and facilitating adaptation to varying environments. While data augmentation techniques have been widely used in S-DG for image classification~\cite{sdgda}, object detection tasks present additional complexities due to the need for accurate detection and classification of multiple objects within a single image. Directly applying data augmentation methods designed for image classification to object detection may significantly compromise object annotations. Several approaches, such as style-transfer augmentation~\cite{michaelis2019benchmarking}, have mitigated this issue by preserving object positions, yet they often fail to fully exploit the range of transformations available in image classification, resulting in limited domain coverage and insufficient generalization.

\begin{figure}[!t]
\centering 
\includegraphics[width=\linewidth]{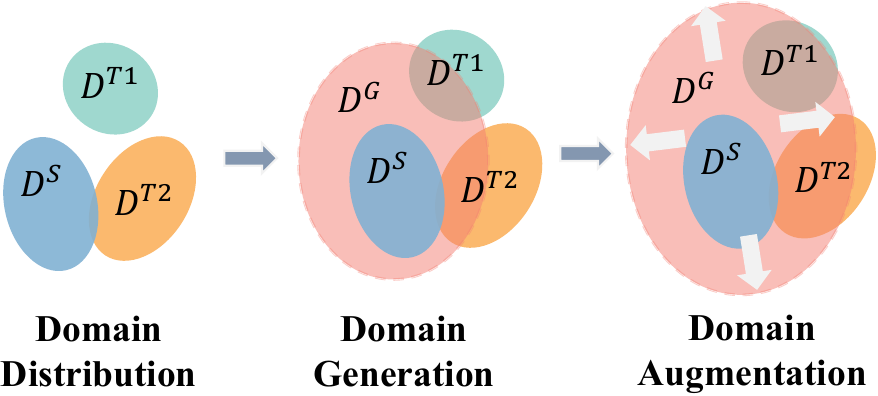}
\caption{
Our method expands the training data distribution through a two-step approach. First, LDMs generate a pseudo-target domain $D^{G}$ with diverse styles, partially covering multiple target domains $D^T$. This $D^{G}$ is combined with the source domain $D^S$ for model training. Second, during training, the data distribution is further augmented to maximize coverage of potential target domains, enhancing the model's generalization capability.} 
\label{overview}
\end{figure}
\begin{figure*}[!t]
\centering 
\includegraphics[width=\linewidth]{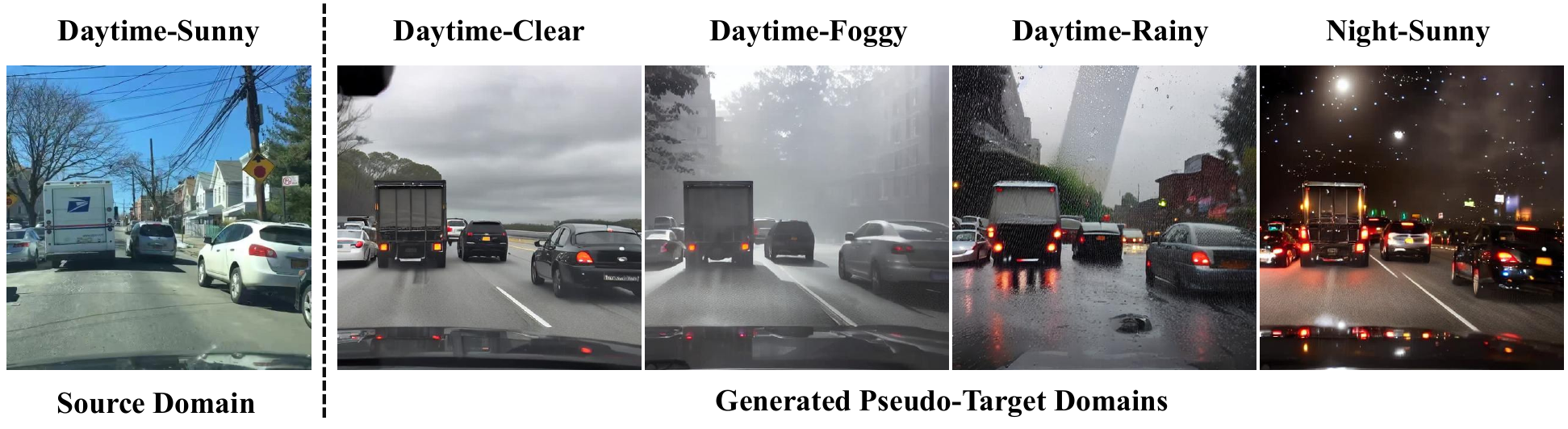}
\caption{Visualization of source domain and generated pseudo-target domains. The leftmost image shows a real Daytime-Sunny scene from the source domain. The four images on the right demonstrate our method's capability to generate diverse pseudo-target domain images while preserving the semantic content and annotations from the source domain.} 
\label{fig:generateimage}
\end{figure*}

Recent advancements in large-scale vision models have highlighted the potential of utilizing the extensive prior knowledge embedded in foundational models pre-trained on internet-scale datasets \cite{schuhmann2022laion}. For instance, the C-Gap model~\cite{ViditES23} achieves State-of-the-Art (SOTA) performance in single-domain object detection by extracting relevant semantic information from the CLIP~\cite{clip} model using domain-specific text prompts. Building on this success, we explore whether the prior knowledge within the Latent Diffusion Models (LDMs)~\cite{ldm}--known for their exceptional performance in image generation--can inform data augmentation strategies to enhance the generalization of object detection tasks in autonomous driving scenarios.

In this paper, we introduce a novel object-style diffusion-based method, called GoDiff, to incorporate extensive transformations at the object level while preserving object annotations for data augmentation. GoDiff integrates a Pseudo Target Data Generation (PTDG) module with a Cross-Style instance Normalization (CSN) technique to perform data augmentation at both image and feature levels. The PTDG module aims to transform an LDM into a pseudo-target domain generator, generating virtual images with varied styles but consistent annotations from a single source domain (see \ref{fig:generateimage}). To achieve this, we propose a dual-prompt strategy for the LDM to effectively diverge the style distribution of objects. Furthermore, considering that the generated virtual images may contain low-quality objects, we introduce an object filtering method using the CLIP-RBF kernel to remove these, ensuring data reliability. Additionally, CSN further extends the scope of data augmentation by exchanging style features across different generated pseudo-domains (using PTDG) and the source domain. This interaction of cross-domain style features enhances model robustness thereby improving generalization ability. By combining dual-level data augmentation techniques at both image and feature levels, our method provides richer training data and more robust feature representations for object detector models, significantly enhancing their generalization performance in unseen scenarios.

Figure 1 illustrates our generalization strategy. We use PTDG to expand the training distribution at the image level, and propose CSN to enhance domains at the feature level. This dual approach improves the model's ability to encompass the target domain distribution, enhancing generalization performance.

The key contributions of this work are as follows:

{\romannumeral 1}) We propose an innovative image generation module PTDG that requires only a single annotated source domain to produce data with consistent annotations across diverse styles.  Integrating these generated datasets significantly enhances the detector's performance, particularly when encountering previously unseen scene variations.

{\romannumeral 2}) We propose a novel feature enhancement technique CSN that performs feature-level style exchange on images from multiple generated pseudo-domains during training. This approach significantly increases the model's robustness and generalization ability.


{\romannumeral 3}) By adopting the proposed two methods, GoDiff significantly enhances the generalization performance of existing object detection models. Notably, as a versatile, plug-and-play tool, GoDiff enhances other S-DG methods, achieving SOTA performance in autonomous driving scenarios.

%% file: sections/Related_Work.tex
\begin{figure*}[!t]
\centering 
\includegraphics[width=7.0in]{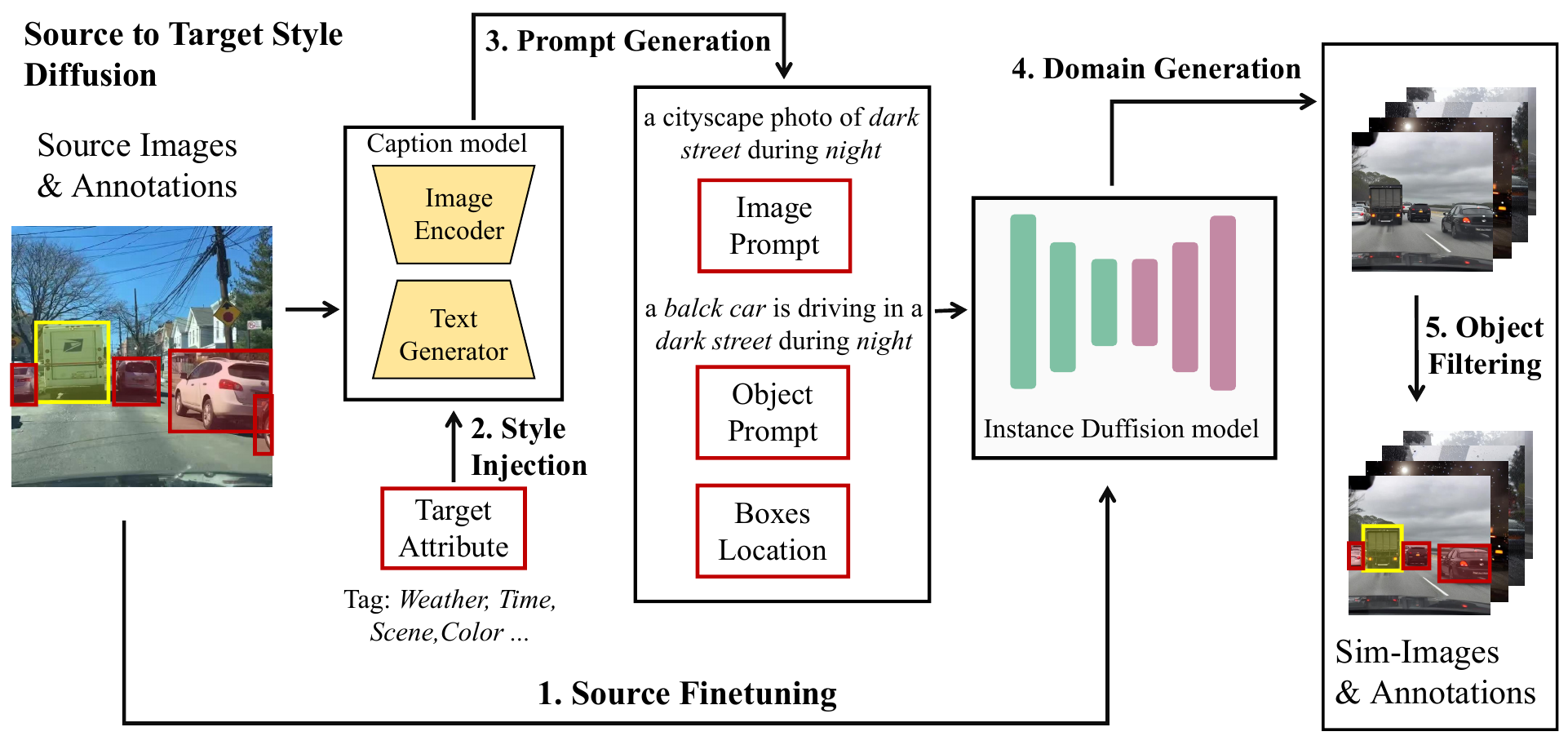}
\caption{ Pipeline for pseudo-target domain generation. The process takes annotated source domain images as input and produces annotated images for multiple pseudo-target domains. This approach enables the creation of diverse, style-rich images while preserving original annotations, facilitating domain generalization in object detection tasks.} 
\label{pipeline}
\end{figure*}

\section{Related Work}

\subsection{Data Augmentation for Domain Generalization}
Data augmentation is a highly effective method for single-domain generalization. In image classification tasks, previous efforts have produced diversified domains through mixed augmentation chains \cite{hendrycks2019augmix} and orthogonal primitive transformations \cite{modas2022prime}. However, these methods may alter the position and semantic features of objects, thereby decreasing the model performance on object detection tasks. To avoid this issue, \cite{geirhos2018imagenet} expands the domain coverage through style transfer \cite{ImageStyleTransfer}, which does not change the position of objects and eliminates texture biases unrelated to the objects, thus enhancing the generalization of the object detection model. However, this limited enhancement method is insufficient, and more exploration is still needed in the domain generalization of object detection.

\subsection{Diffusion Models for Domain Generalization}

Diffusion Models (DMs) \cite{dm1,dm2} have shown exceptional potential in deep generative models, outperforming GANs and VAEs in image generation quality. LDMs \cite{ldm} have further improved efficiency by operating in latent space and incorporating text-guided techniques like CLIP \cite{clip}. Recent research has explored DMs for generalizing computer vision tasks \cite{jia2023dginstyle,segdg}. (\citeauthor{gong2023} 2023) proposed prompt randomization to enhance cross-domain performance, while DatasetDM \cite{wu2023datasetdm} introduced a universal dataset generation model. These approaches primarily address semantic segmentation tasks. However, applying diffusion models to domain generalization in object detection remains understudied. Unlike semantic segmentation, object detection requires both semantic consistency and precise object-level alignment. The recent introduction of InstanceDiffusion \cite{wang2024instancediffusion}, which provides instance-level control in text-to-image diffusion models, offers potential solutions for creating object-aligned datasets suitable for detection tasks.

\subsection{Single-domain generalization for object detection}


Although there are numerous single-domain generalization methods for image classification, these methods cannot be directly applied to object detection tasks as they disrupt the features of the objects. To address this, \cite{oadg} proposed an object-aware domain generalization approach, and \cite{ViditES23} introduced a novel semantic enhancement strategy using a pre-trained vision-language model. In addition, the Generalizable Neural Architecture Search (G-NAS) \cite{gnas} was proposed to adapt to task complexity by altering network capacity, preventing small-capacity networks from overfitting easy-to-learn, irrelevant features like background characteristics. Cyclic Separation Self-Distillation \cite{singleDGOD} learns domain-invariant object features unsupervisedly, demonstrating good generalizability and further enhancement via self-distillation.

%% file: sections/Method.tex
\section{Method}

\subsection{Approach Overview}



Single-domain generalization for object detection aims to enable models to perform well in unseen domains using annotated data from a single source domain. To address this challenge, we propose a novel object-style diffusion-based method called GoDiff for data augmentation. 
Our GoDiff integrates a PTDG module with a CSN technique to perform data augmentation at both image and feature levels. Specifically, the PTDG module synthesizes diverse datasets of labeled virtual images (see Fig.~\ref{pipeline}). Starting with the annotated source domain dataset, our method utilizes an instance diffusion model to generate pseudo target datasets. This process involves generating global and instance-level textual prompts for each source image, guiding the diffusion model to produce stylistically diverse virtual images while maintaining the semantic integrity of object positions and labels. The resulting pseudo-target dataset domain dataset is then combined with the original source domain dataset to form an augmented training set.

Given the potential distribution bias between domains in the augmented dataset, we propose a CSN training strategy to facilitate smooth transitions between different styles. CSN broadens the training distribution by mixing image features of various styles at multiple network levels. The generated data retains the same annotations as the real data, and the high stylistic diversity within the generated domain enhances the training dataset's distribution. By combining real and virtual data during training, our approach maximizes the model's generalization ability.

\subsection{Object Conditioned Image Generation}

Recent advances in image generation, particularly through latent diffusion models like Stable Diffusion \cite{ldm}, have opened new avenues for creating synthetic training data for neural networks. While previous work has successfully applied these techniques to semantic segmentation tasks, adapting them for object detection presents unique challenges. Our approach addresses two key aspects of image generation for object detection: scene diversity and object-level control. We utilize InstanceDiffusion \cite{jia2023dginstyle} as our core generative model and fine-tune it on our source domain dataset to better capture the nuances of street scenes. We also incorporate a CLIP text encoder and Fourier feature mapping to create embeddings for each object instance. After training, we constrain the generated objects to the bounding box regions specified in the source domain labels. This approach allows us to generate diverse, yet structurally consistent, synthetic data for improving object detection models.

\subsection{Dual-prompt Generation Strategy}

The generation of diverse and realistic pseudo-target domains requires object-level descriptions and conditions to guide image creation. However, no established solution exists for designing object-level text prompts for diffusion models in complex urban street scenes. To address this, we developed a dual-prompt generation strategy that controls overall scene characteristics with image-level prompts and fine-tunes individual object attributes using object-level prompts. This approach enhances dataset diversity, potentially improving the robustness and generalization of object detection models trained on it.

\subsubsection{Image-level Prompt.}
The image-level prompt describes global image features, including weather, time, and scene. A caption model automatically extracts key global features from source domain images and combines them with target domain descriptors to generate natural language descriptions, reflecting the overall style and context.

We leverage the Tag2Text \cite{huang2023tag2text} to extract descriptive tags from source domain images and augment them with target domain descriptors. This process is as follows:

\textbf{Tag Extraction}. Let $X_s$ be a source domain image. We utilize the Tag2Text to extract a list of tags $t_{si}$, represented as:
\begin{equation}
\mathcal{T}_s=Tag(X_s)=\{t_{s1},t_{s2},\ldots,t_{sn}\},
\end{equation}
where $\mathcal{T}_s$ is the set of tags extracted from $X_s$.

\textbf{Tag Augmentation}. We introduce a set of domain descriptors $ \mathcal{T}_d = \{t_{d1}, t_{d2}, \ldots, t_{dm}\}$, including weather, time, and specific scenes. The augmented tag set $\mathcal{T}_{aug}=\mathcal{T}_s \cup \mathcal{T}_d$, where $\cup$ denotes the union of source and target tag sets.

\textbf{Prompt Construction}. The combined tag set $\mathcal{T}_a$ is synthesized into a descriptive sentence ($P_w$) using a text decoder $Dec$:
\begin{equation}
P_w=Dec(\mathcal{T}_{aug}).
\end{equation}

For example, if $\mathcal{T}_s$ = {\text{``cityscapes, street"}} and $\mathcal{T}_d$ = {\text{"night, dark"}}, $P_w$ could be: ``a cityscapes photo of a dark street during night."

\subsubsection{Object-Level Prompt.}

object-level prompts detail specific objects in an image with attributes like actions, positions, colors, etc. Unlike automated image-level prompts, the random combination approach generates numerous distinct instance prompts through predefined templates filled with specific content, offering greater flexibility.

We define a template $\mathcal{P}$ as a structured string containing placeholders for descriptors such as object, action, weather, scene, and time. This template is expressed as: $\mathcal{P}=\text{``A \textit{object} is \textit{action} in a \textit{weather} \textit{scene} during \textit{time}."}$ Let $o$, $a$, $w$, $s$, and $t$ represent placeholders for object, action, weather, scene, and time, respectively. Each descriptor category has its own set: $\mathbf{O}=\{o_1,o_2,\ldots,o_x\},\quad\mathbf{A}=\{a_1,a_2,\ldots,a_y\},\quad\mathbf{W}=\{w_1,w_2,\ldots,w_z\},\quad\mathbf{S}=\{s_1,s_2,\ldots,s_u\},\quad\mathbf{T}=\{t_1,t_2,\ldots,t_v\}$. $c$ must be consistent with the real category, but modifiers can be added, e.g., the category ``car" can be adjusted to ``yellow taxi".

An object-level prompt, denoted as $P_i$, is generated by randomly selecting elements from each set:
\begin{equation}
P_i=\mathcal{T}(o_x,a_y,w_z,s_u,t_v),
\end{equation}
where $o_x \in \mathbf{O}$, $a_y \in \mathbf{A}$, $w_z \in \mathbf{W}$, $s_u \in \mathbf{S}$, and $t_v \in \mathbf{T}$ are randomly chosen descriptors. For example, a generated prompt might be ``A black car is parked in a foggy street during the night."

\subsection{Pseudo-Target Domain Generation}

We employ our dual-prompt generation scheme with object location information from source domain images to generate diverse virtual images, expanding the source domain into multiple pseudo-target domains.

Given an annotated source image $\mathbf{X}_s$ with object locations $\mathbf{B}_s = \{b_1, b_2, \ldots, b_n\}$, we create a global textual description $P_w$ of the target domain scene and individual style descriptions $p_i$ for each bounding box region. The source image $\mathbf{X}_s$, global description $P_w$, and instance-specific conditions $\mathbf{C} = \{(p_1, b_1), (p_2, b_2), \ldots, (p_n, b_n)\}$ are input into a customized InstanceDiffusion model, generating a virtual image $\mathbf{X}_t$:
\begin{equation}
\mathbf{X}_{t}=\mathcal{F}\left(\mathbf{X}_{s}, P_w, \mathbf{C} ; \theta\right).
\end{equation}

The generator $\mathcal{F}$ starts with a noise image $\mathbf{z} \sim \mathcal{N}(\mathbf{0}, \mathbf{I})$ and produces the final virtual sample through denoising:
\begin{equation}
    \mathbf{X}_{t}=\mathbf{z}-\sum_{i=1}^{T} \mathcal{F}\left(\mathbf{z}_{i}, P_w, \mathbf{C} ; \theta\right).
\end{equation}

InstanceDiffusion integrates instance-specific conditioning into visual features, and generated images exhibit the target style at specified locations. Annotations $\{\mathbf{B}_s, \mathbf{Y}_s\}$ from $\mathbf{X}_s$ are transferred to $\mathbf{X}_t$, creating a virtual sample $(\mathbf{X}_t, \mathbf{B}_s, \mathbf{Y}_s)$ while preserving semantic consistency.

By generating multiple stylized images per source image, we obtain pseudo-target domain datasets $ \mathcal{D}^{G_{k}}$. These, combined with the source domain dataset $ \mathcal{D}^S$, form an expanded training set enhancing the object detection model's domain generalization ability.

\begin{figure}[!t]
\centering 
\includegraphics[width=\linewidth]{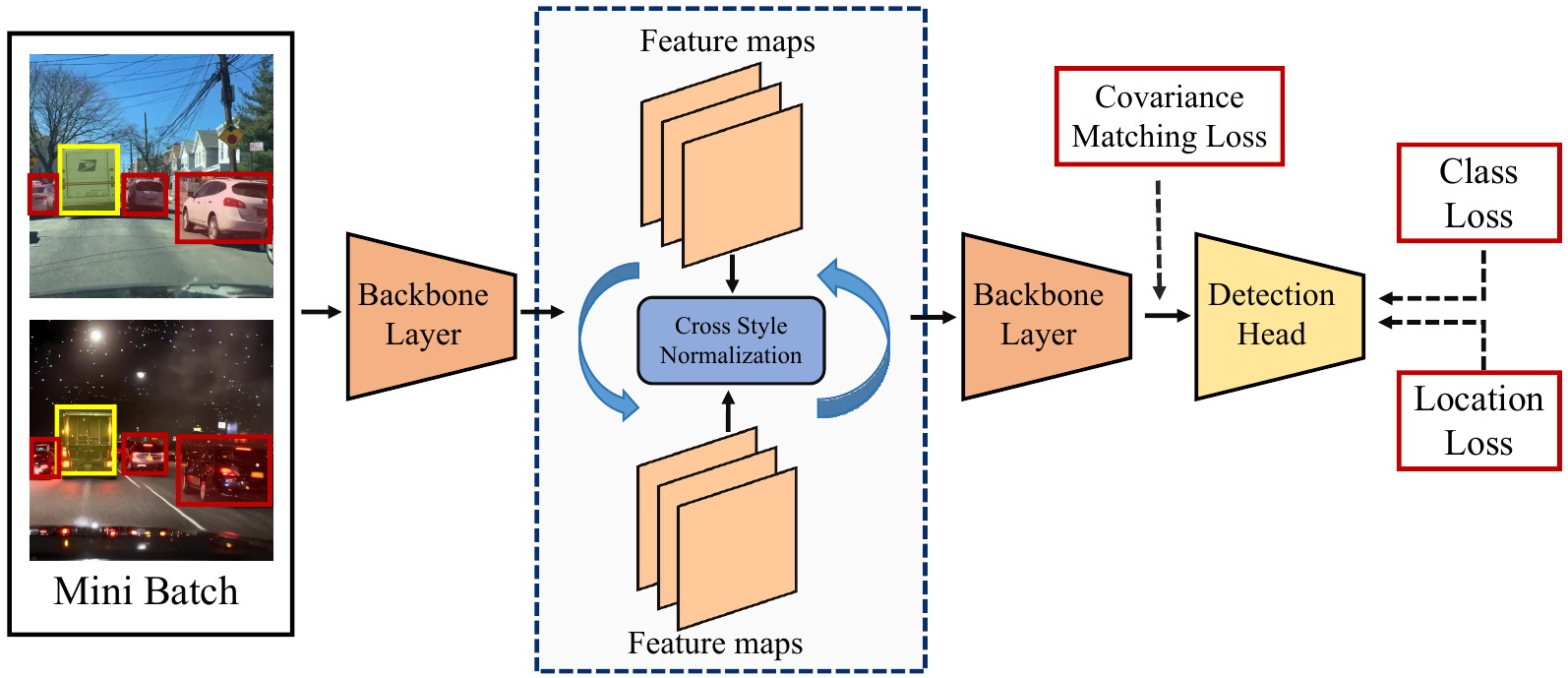}
\caption{Cross Style Normalization-based detector training framework.  The CSN modules are embedded between Backbone layers, processing feature maps from images with diverse styles. This approach leverages diverse image styles to create a more continuous feature distribution, promoting style-invariant feature learning for improved domain generalization.} 
\label{CSN}
\end{figure}

\subsection{Object Filter}

In generating pseudo-target domain images, object quality is crucial for accurate alignment. Poor-quality objects can lead to inconsistencies, necessitating quality detection and filtering. CMMD \cite{cmmd} with a Gaussian RBF kernel \cite{rbf,kernel2, gretton2012kernel}accurately measures discrepancies between generated and real image distributions. However, it primarily focuses on distribution differences. Inspired by CMMD, we propose an object filtering mechanism using the CLIP-RBF kernel. This approach evaluates semantic disparities between generated and real images, ensuring elimination of inconsistent objects and improving dataset quality.

For a given source domain image $(\mathbf{X}_s, \mathbf{B}_s, \mathbf{Y}_s)$, we generate a corresponding virtual source domain image $(\mathbf{X}_{PS}, \mathbf{B}_s, \mathbf{Y}_s)$, where both images depict the same scene, such as a Daytime-Sunny setting. Here, $\mathbf{B}_s = \{b_1, b_2, \dots, b_n\}$ represents the set of bounding box coordinates within the image. We assume that the objects in these two images should maintain a high degree of semantic consistency.

Building on this, for each bounding box $b_i \in \mathbf{B}_S$, we extract the features from both the source domain image and the virtual source domain image within that bounding box. Using the CLIP image encoder, we encode these region features into feature vectors of the same dimension, denoted as $\mathbf{f}_S(b_i)$ and $\mathbf{f}_{PS}(b_i)$, respectively. The similarity between these vectors is then measured using the CLIP-RBF distance, as follows:
\begin{equation}
\small
    D_\mathrm{RBF}(\mathbf{f}_S(b_i),\mathbf{f}_{PS}(b_i))=\exp\left(-\gamma\|\mathbf{f}_S(b_i)-\mathbf{f}_{PS}(b_i)\|^2\right),
\end{equation}
where $\gamma$ is a hyperparameter that controls the spread of the RBF kernel. If the distance $D_\mathrm{RBF}$ exceeds a predetermined threshold $\tau$, the corresponding bounding box is filtered out; otherwise, the bounding box $b_i$ is retained. This process ensures that only bounding boxes with highly similar features between the source and virtual source domains are preserved. The resulting set of retained bounding boxes is denoted as $\mathbf{B}_{\hat{S}}$:
\begin{equation}
    \mathbf{B}_{\hat{S}}=\{b_i\in\mathbf{B}_S\mid D_{\mathrm{RBF}}(\mathbf{f}_S(b_i),\mathbf{f}_{PS}(b_i))\leq\tau\}.
\end{equation}

The filtered bounding boxes, along with their corresponding object categories $(\mathbf{Y}_{\hat{S}}, \mathbf{B}_{\hat{S}})$, serve as labels for all pseudo-target domains. This approach ensures that the generated pseudo-labels are more accurate and consistent, thereby improving the model's performance on the target task.

\input{tables/DAD}

\subsection{Cross Style Normalization}

To effectively integrate the generated pseudo target domains with the original domain data, we developed a strategy to address domain discrepancies and distribution biases, aiming for a coherent and diverse training distribution.

Ulyanov et al. introduced Instance Normalization \cite{ulyanov2016instancenorm}, primarily enhancing style transfer performance. Unlike Batch Normalization, Instance Normalization uniquely normalizes each channel per sample, expressed as:
\begin{equation}
   \widetilde{F} = \mathrm{IN}(F) = \gamma\left(\frac{F - \mu(F)}{\sigma(F)}\right) + \beta,
\end{equation}
where $F$ is the input, and $\gamma$, $\beta$ are scaling and shifting parameters. The functions $\mu(\cdot)$ and $\sigma(\cdot)$ represent the mean and standard deviation, encoding style elements such as color and contrast. Thus, the style of a feature map $Z \in \mathbb{R}^{C \times H \times W}$ is defined by channel-wise $\mu(Z), \sigma(Z) \in \mathbb{R}^C$, calculated as follows:
\begin{equation}
\small
\mu(\mathbf{Z}) = \frac{1}{HW} \sum_{h=1}^{H} \sum_{w=1}^{W} \mathbf{Z}_{\cdot,h,w},
\end{equation}

\begin{equation}
\small
\sigma(\mathbf{Z}) = \sqrt{\frac{1}{HW} \sum_{h=1}^{H} \sum_{w=1}^{W} \left(\mathbf{Z}_{\cdot,h,w} - \mu(\mathbf{Z})\right)^{2}}.
\end{equation}

Inspired by Instance Normalization \cite{ulyanov2016instancenorm}, which normalizes each channel of each sample independently, we propose Cross Style Normalization (CSN). This strategy enables the mixing of image features across different styles at various levels of the network, thereby broadening the training distribution. 

The CSN operation involves dividing images from the source and pseudo-target domains into small batches, extracting feature maps, and then randomly exchanging the mean and variance of these maps between image pairs. For feature maps $F_A$ and $F_B$ of the image pair $(X_A, X_B)$, the transformation is defined as:
\begin{equation}
\small
\widetilde{F}_A = \sigma(F_B)\frac{F_A-\mu(F_A)}{\sigma(F_A)}+\mu(F_B)
\end{equation}
\begin{equation}
\small
\widetilde{F}_B = \sigma(F_A)\frac{F_B-\mu(F_B)}{\sigma(F_B)}+\mu(F_A)
\end{equation}

This style-swapping mechanism, triggered with a certain probability during training, efficiently augments styles and expands the sample distribution.

Additionally, we introduced a Covariance Matching Loss (CML) \cite{cml} to minimize differences between the covariance matrices of features from different styles, ensuring the learning of style-invariant features. Flattening $\widetilde{F}_A$, $\widetilde{F}_B$ into $\bar{F}_{A}$ and $\bar{F}_{B}\in\mathbb{R}^{(HW)\times C}$, the covariance matrices are expressed as:
\begin{equation}
\small
\Sigma_{X_A,X_A}^i=\bar{F_A^i}^T\cdot\bar{F_A^i},\quad \Sigma_{X_B,X_B}^i=\bar{F_B^i}^T\cdot\bar{F_B^i}.
\end{equation}
CML is defined as:
\begin{equation}
\small
    \mathcal{L}_{CM}=\|\Sigma_{X_A,X_A}^i-\Sigma_{X_B,X_B}^i\|_2.
\end{equation}

This approach allows the network to form a more continuous and diverse feature distribution, enabling the detector to learn style-invariant features from a broader and more varied sample set.

%% file: tables/DAD.tex
\newcommand{\mycheckmark}{\ding{52}}
\newcommand{\mycrossmark}{\ding{56}}
\definecolor{ForestGreen}{RGB}{34,139,34}

\begin{table*}[t]
        \makeatletter\def\@captype{table}\makeatother
        \centering
        \caption{Comparison with state-of-the-art methods on different weather conditions. 
        }
        \centering
        \addtolength{\tabcolsep}{3pt}
        \begin{tabular}{l|c|c|cccc|cc}
        \shline
        \multicolumn{1}{l|}{\multirow{1}{*}{Method}} &\makecell{GoDiff} & \makecell{Daytime \\  Sunny}  & \makecell{Night \\ Sunny} & \makecell{Dusk \\ Rainy} & \makecell{Night \\ Rainy} & \makecell{Daytime \\ Foggy} & \makecell{mPC (\%)$\uparrow$} & \makecell{relative\\gain (\%)$\uparrow$} \\ \midrule
        F-RCNN & \ding{56} & 50.2 &   31.8 & 26.0 & 12.1 & 32.0   & 25.5   & -    \\

        SW & \ding{56} & 50.6 & 33.4 & 26.3 & 13.7 & 30.8 & 26.1   &   0.6    \\
        IBN-Net & \ding{56}  & 49.7 & 32.1 & 26.1 & 14.3 & 29.6 & 25.5 & 0.0        \\
        IterNorm & \ding{56}  & 43.9 & 29.6 & 22.8 & 12.6 & 28.4  & 23.4    &  -2.1    \\
        ISW& \ding{56} & 51.3 & 33.2 & 25.9 & 14.1 & 31.8 &  26.0 & 0.5 \\
        Ours & \ding{52} & \textbf{55.0} & \textbf{35.4} & \textbf{32.1} & \textbf{15.0} & \textbf{35.9} & \textbf{29.6} & \textbf{4.1}\\ \midrule
        S-DGOD & \ding{56} & 56.1 & 36.6 & 28.2 & 16.6 & 33.5 & 28.7  & 3.2 \\
        C-Gap& \ding{56} & 48.1 & 32.0  & 26.0 & 12.4 & 34.4 & 26.2 & 0.7
 \\ 
        SRCD& \ding{56} &- & 36.7  &28.8  &17.0  &35.9  &29.6 & 4.1  \\
        OA-DG & \ding{56} & 55.8 & 38.0 & 33.9 & 16.8 & 38.3 & \textbf{31.8} &\textbf{6.3}\\ 

        OA-DG& \ding{52} & \textbf{56.0} & \textbf{38.4} & \textbf{34.1} & \textbf{18.0} & \textbf{39.9} & \textbf{32.6} & \textbf{7.1}
 \\         \shline
        \end{tabular}
        \label{table:dwd}
        \vspace{-2pt}
    \end{table*}


%% file: sections/Expriments.tex
\section{Experiment}

\subsubsection{Datasets and Metrics}

We employ two essential datasets for our study: 1)~The Diverse Weather Dataset (DWD) \cite{singleDGOD}, which contains urban scenes under five distinct weather conditions; and 2)~The Cityscapes-C \cite{city-c} dataset, a benchmark to evaluate model robustness against 15 different types of corruption. We use the mean Average Precision (mAP) and mean Performance under Corruption (mPC) assess domain generalization and robustness against out-of-distribution data \cite{oadg}.

\subsubsection{Implementation Details}
We employ InstanceDiffusion \cite{ldm}, fine-tuned on the Daytime-Sunny dataset, as the base image generation model. We employ the Tag2Text \cite{huang2023tag2text} to generate text prompts. Our object detector model is based on Faster R-CNN \cite{fasterrcnn} with ResNet101 backbone, initialized with ImageNet weights. The training is conducted on the Daytime-Sunny dataset and generated pseudo-target domains, including Night-Sunny, Night-Rainy, Daytime-Foggy, and Dusk-Rainy. We use the SGD with momentum of 0.9 and weight decay of 1e--4 as the optimizer. The learning rate is set to 1e--2, with a batch size of 4 per GPU. Further details on the experimental setup and model configurations are provided in the \textbf{supplementary materials}.

\subsection{Comparison Results}

\textbf{Robustness Across Various Weather Conditions}. We present a comprehensive comparison of our method against eight SOTA approaches, including SW \cite{SW}, IBN-Net \cite{IBN-Net}, IterNorm \cite{IterNorm}, ISW \cite{ISW}, and recent SDGOD methods such as S-DGOD \cite{singleDGOD}, C-Gap \cite{ViditES23}, SRCD \cite{rao2023srcd} and OA-DG \cite{oadg}, under various weather conditions in real-world urban environments. 

As shown in Table~\ref{table:dwd}, Our method consistently outperforms others across all target domains, with particularly notable results in daytime-sunny and dusk-rainy conditions, achieving scores of 55.0 and 32.1, respectively—significantly higher than those of other methods. GoDiff also shows strong performance in the night-sunny and dusk domains, with scores of 35.4 and 32.1. Furthermore, we extend GoDiff to the OA-DG framework, and the results demonstrate that our generated dataset significantly enhances performance in most target domains, setting new SOTA benchmarks. These results underscore the effectiveness of our PTDG pipeline.

\textbf{Robustness Against Common Corruptions}. We present a comprehensive comparison of our method against two SOTA approaches (OA-Mix and OA-DG \cite{oadg})  and seven conventional data augmentation techniques (Cutout, Photo, AutoAug-det, AugMix, Stylized, SupCon and FSCE) \cite{oadg}, under various types of common corruptions. All models are trained on clean domain data and then evaluated across both clean and corrupted domains. 

Table~\ref{tab:cityc} presents the effectiveness of each method. As shown, conventional augmentation techniques yield minimal improvements in corrupted domains, whereas our GoDiff method demonstrates significant gains. However, GoDiff is less effective against digital corruption types, likely due to the introduction of noise and distortion patterns that differ from those in natural images. Nevertheless, when integrated with OA-DG, GoDiff significantly improves performance, establishing new SOTA benchmarks. This synergy underscores the potential of generative data in enhancing model robustness against common corruptions.

\input{tables/cityscape_c}

\subsection{Ablation Study}

We conduct comprehensive ablation studies to assess the contribution of each component in our proposed method and present the results in Table \ref{tab:ablation_study}. 
Firstly, the inclusion of Pseudo-Target domain training (PT) significantly improves performance across various weather conditions, particularly in Day Foggy, where mAP increases from 32.0 to 33.7. This indicates that effectively utilizing generated pseudo-target domain data enhances domain generalization.
Secondly, combining CSN with PT further improves performance, notably in Night Sunny (mAP from 29.0 to 33.9) and Daytime Foggy (mAP from 32.4 to 35.2). This integration aligns feature distributions across various styles, enhancing robustness.
Thirdly, the addition of CML  achieves the best overall performance, with a further improvement of +1.5 mAP in Night Sunny. This explicitly aligns with inter-domain covariance statistics, demonstrating its effectiveness.
Fourthly, omitting any component results in a performance decline, affirming the value of each component. In challenging scenarios like Night Rainy, the complete model achieves an mAP of 15.0, significantly improving over the baseline's mAP of 12.1.
These findings validate the design rationale of our method and underscore the importance of each component in achieving robust object detection across diverse weather conditions.

\subsection{Additional Experimental Analysis}






\textbf{Versatility Across Different Detectors}. We apply GoDiff to several object detection architectures, including Faster R-CNN, Mask R-CNN, and DINO, to demonstrate its versatility. As the results presented in Table~\ref{tab:detector}, GoDiff shows a consistent performance improvement across different detector backbones and weather conditions and enhances the mPC for all different object detectors. Specifically, Faster R-CNN shows a 4.3-point increase in mPC, with significant improvements in challenging scenarios like Night Rainy (from 12.1 to 15.0) and Dusk Rainy (from 26.0 to 32.1). Mask R-CNN shows a 4.1-point increase in mPC, with notable enhancements in Night Sunny (from 30.3 to 37.2) and Dusk Rainy (from 27.1 to 30.4) conditions. As a SOTA detector, DINO shows the most substantial improvement with a 4.7-point increase in mPC, particularly in adverse conditions such as Night Rainy (from 8.3 to 12.4) and Day Foggy (from 33.3 to 38.8).
These results demonstrate GoDiff's generalizability and its ability to consistently enhance performance across various state-of-the-art object detection models, especially in challenging weather conditions. 

For further experimental results, visualizations, and technical details, please refer to the \textbf{Supplementary Materials}.

\input{tables/Ablation}

\input{tables/detector_DAD}

%% file: tables/cityscape_c.tex
\definecolor{ForestGreen}{RGB}{34,139,34}

\begin{table}[t]
        \makeatletter\def\@captype{table}\makeatother
        \centering
        \caption{Comparison with state-of-the-art methods on Cityscapes-C. For each corruption type, average performance was calculated. mPC denotes the average performance across 15 corruption types. $\dagger$ indicates that the method is enhanced by our GoDiff implementation.} \label{tab:cityc}
        \vspace{1.6mm}
        \centering
        \addtolength{\tabcolsep}{-3pt}
        \begin{tabular}{l|c|cccc|c}
        \shline
        \multicolumn{1}{l|}{\multirow{1}{*}{Method}} &\makecell{Clean} &\makecell{Noise} & \makecell{Blur}  & \makecell{Weather} & \makecell{Digital} & \makecell{mPC} \\ \midrule
        F-RCNN & 42.2 &  0.9   & 13.5  &  19.1 &  24.5  & 15.4       \\ \midrule
        
        Cutout  & 42.5 & 1.0  & 13.7  & 19.6  & 24.8  & 15.7          \\
        Photo & 42.7  & 2.1  & 13.2 & 23.5  & 25.3 & 16.9        \\
        AutoAug-det & 42.4  &  1.1 & 13.0  & 21.5  & 23.8  & 15.8        \\
        AugMix & 39.5  & 5.7  & 15.5  &  22.6 & 25.7  & 18.1  \\
        Stylized & 36.3  & 5.3  & 14.9  & 20.7  & 25.0  & 17.2        \\
        OA-Mix & 42.7  & 8.2  &  17.2 &  26.2 & 28.6  & 20.8        \\
        SupCon & 43.2  &  8.0 & 17.6  &  26.1 & 29.0  & 20.9        \\
        FSCE & 43.1  &   8.6 & 17.5  & 26.1  & 28.9  & 21.0        \\
        OA-DG & 43.4  &  9.1  &  18.1 &  \textbf{27.3} &  \textbf{29.7} & 21.8        \\ \midrule
        
        GoDiff  & 42.7  & 8.8  &  17.9 & 27.5  & 26.8  & 20.3 \\ 
        OA-DG~$\dagger$  & 42.9  &  \textbf{10.9}  &  \textbf{18.2} & 27.1  & 29.4  &  \textbf{22.8}   \\
         \shline
        \end{tabular}
        \label{table:dwd}
        \vspace{-2pt}
    \end{table}

%% file: tables/Ablation.tex


\begin{table}[t]
    \centering
        \caption{
Ablation study results for Faster R-CNN on DWD dataset. PT: Pseudo Target domain training, CSN: Cross-Style Normalization, CML: Covariance Matching Loss.
    }
    \begin{adjustbox}{width=\linewidth}
        \begin{tabular}{@{}cccc|cccccc@{}}
            \shline
            \textbf{PT} & \textbf{CSN} & CML  & \makecell{Day\\Sunny}& \makecell{Night\\Sunny} & \makecell{Dusk\\Rainy} &\makecell{Night\\Rainy}   & \makecell{Day\\Foggy} \\
            \midrule
            \ding{56} & \ding{56} & \ding{56} & 50.2 & 31.8 & 26.0 & 12.1 & 32.0 \\
            \ding{52} & \ding{56} & \ding{56} & 51.9 & 32.0 & 28.4 & 12.5 & 33.7 \\
            \ding{56} & \ding{52} & \ding{56} & 53.5 & 29.0 & \textbf{32.4} & 13.2 & 32.4 \\
            \ding{52} & \ding{52} & \ding{56} & \textbf{55.4} & \underline{33.9} & 32.0 & \underline{14.6} & \underline{35.2} \\
            \ding{52} & \ding{52} & \ding{52} & \underline{55.0} & \textbf{35.4} & \underline{32.1} & \textbf{15.0} & \textbf{35.9} \\
            \shline
        \end{tabular}
    \end{adjustbox}

    \label{tab:ablation_study}
\end{table}

%% file: tables/detector_DAD.tex

\begin{table}[t!]
\centering
\caption{ Performance comparison of different object detectors with and without GoDiff on the DWD dataset.
}
\label{tab:detector}
\setlength{\tabcolsep}{4pt}
\resizebox{1.0\columnwidth}{!}{
\begin{tabular}{@{}l c ccccc cc @{}}

\shline
\addlinespace[5pt]

  Detector &
  GODiff &
  \makecell{Day\\Sunny} &
  \makecell{Night\\Sunny} &
  \makecell{Dusk\\Rainy} &
  \makecell{Night\\Rainy} & 
  \makecell{Day\\Foggy} & 
  mPC  \\

\addlinespace
\midrule

\multirow{3}{*}{F-RCNN}& 
\ding{56}      & 50.2 & 31.8 & 26.0 & 12.1 & 32.0 &\multirow{2}{*}{{$\uparrow$ \textbf{4.3}}}\\
 & \ding{52}   & \textbf{55.0} & \textbf{35.4} & \textbf{32.1} & \textbf{15.0} & \textbf{35.9} \\ 
\midrule

\multirow{3}{*}{M-RCNN}  
& \ding{56} & 53.2 & 30.3 & 27.1 & 11.4 & 34.1 &\multirow{2}{*}{{$\uparrow$ \textbf{4.1}}} \\
& \ding{52} &\textbf{58.4}  & \textbf{37.2}  & \textbf{30.4}  & \textbf{13.9} & \textbf{36.7}   \\
\midrule


\multirow{3}{*}{DINO} 
& \ding{56} & 53.8 & \textbf{29.7}  & 24.6  & 8.3  & 33.3 &\multirow{2}{*}{{$\uparrow$ \textbf{4.7}}}\\
& \ding{52}  & \textbf{58.8} & \textbf{34.5}  & \textbf{30.2}   & \textbf{12.4} & \textbf{38.8}    \\
\shline

\end{tabular}
}
\end{table}

%% file: sections/Conclusion.tex
\section{Conlusion}
In this study, we propose GoDiff, a novel method to improve object detection generalization in urban scenes. By integrating the PTDG module and CSN technique, GoDiff enriches training data and strengthens feature representations, leading to more robust object detection models. Extensive experimental results demonstrate that GoDiff not only enhances the generalization ability of object detector models but also functions as a plug-and-play enhancement for other S-DG methods, achieving SOTA performance in autonomous driving scenarios.

%% file: sections/Appendix.tex
\appendix
\section{Appendix}
\subsection{More Implementation Details}

The CSN module is integrated into the backbone of baseline detector to enhance robustness by exchanging style features across different domains. It is activated with a 0.1 probability in each backbone layer, with a maximum of two layers activated per forward pass. At the core of PTDG module lies the InstanceDiffusion model \cite{wang2024instancediffusion}. We develop this model by fine-tuning Stable Diffusion 1.5 \cite{ldm} on the Daytime-Sunny training set for 40 epochs, adapting it to urban scenes. The fine-tuning process adopt default training configurations, with the exception that the SDXL-refiner is not utilized.

To promote transparency and facilitate further research in this field, we have made our code available at https://anonymous.4open.science/r/GoDiff-FB85/. Furthermore, upon acceptance of this paper, we will publicly release both the code and the generated dataset used in our experiments, ensuring full reproducibility of our results.

\subsection{Further Experiments}

\subsubsection{Comparison of the class-wise improvement}

To further assess the effectiveness of our GoDiff method, we perform a comprehensive analysis of its performance across various object categories and target domains (see Table \ref{tab:classwise}), benchmark it against two baseline methods: S-DGOD \cite{singleDGOD} and SRCD \cite{rao2023srcd}.

GoDiff consistently outperforms these baselines across all target domains and object categories. In the Daytime-Foggy scenario, GoDiff achieves significant improvements, particularly with challenging objects like cars, where it delivers a 5.9\% increase in mAP, surpassing both S-DGOD and SRCD. In the Night-Sunny domain, despite the inherent low-light challenges, GoDiff demonstrates robust performance, with notable gains across all categories, including an 8.3\% increase in mAP for bikes and a 5.1\% increase for persons. These results highlight GoDiff's resilience in nighttime conditions, a crucial factor for continuous autonomous driving operations. The most significant advantage of GoDiff is seen in the Dusk-Rainy domain, characterized by low light and severe weather conditions. In this difficult environment, GoDiff markedly outperforms S-DGOD and SRCD, achieving remarkable improvements, such as a 9.8\% increase in mAP for buses and a 7.8\% increase for trucks. This underscores GoDiff's exceptional capability to manage complex environmental challenges that typically hinder object detection systems.

In contrast to S-DGOD, which exhibits negative gains in some categories within certain domains, GoDiff consistently delivers positive enhancements across all categories and domains. This reliability ensures consistent performance improvements regardless of environmental conditions or object types. Compared to SRCD, GoDiff distinctly excels, particularly in more demanding domains like Dusk-Rainy.

\begin{table}[t]
    \centering
        \caption{
Impact of different generation domains on model performance (mAP).
    }
    \begin{adjustbox}{width=\linewidth}
        \begin{tabular}{@{}c|ccccccccc@{}}
            \shline
            \textbf{Generated Domain} & \makecell{Day\\Sunny}& \makecell{Night\\Sunny} & \makecell{Dusk\\Rainy} &\makecell{Night\\Rainy}   & \makecell{Day\\Foggy} \\
            \midrule
            /   &  50.2 &   31.8 & 26.0 & 12.1 & 32.0  \\
            Daytime-Sunny  & \underline{53.1} & 30.7 & 28.3 & 12.5 & 33.7 \\
            Daytime-Foggy  & 52.2 & 31.8 & 26.0 & 12.1 & \underline{35.0} \\
            Dusk-Rainy  & 53.7 & \underline{33.3} & \textbf{32.6} & 13.9 & 33.7 \\
            Night-Rainy  & 52.1 & 32.0 & 28.3 & 12.5 & 33.9 \\
            Night-Sunny  & 52.5 & 32.6 & 31.0 & \textbf{14.6} & 33.7 \\
            All  & \textbf{55.2} & \textbf{34.9} & \underline{31.7} & \underline{13.7} & \textbf{35.2} \\
            \shline
        \end{tabular}
    \end{adjustbox}

    \label{tab:domain}
\end{table}

\subsubsection{Ablation on Generated Domain}
\begin{table*}[t] \centering
    \caption{Comparison of mAP gain variations across various categories on DWD datasets.}
    \label{tab:classwise}
    \resizebox{\textwidth}{!}{
    \Huge
        \begin{tabular}{c|*{7}{c}|*{7}{c}|*{7}{c}|*{7}{c}}
        \shline
        & \multicolumn{7}{c}{Daytime-Foggy} 
                               & \multicolumn{7}{c}{Night-Sunny} & \multicolumn{7} {c}{Dusk-Rainy} & \multicolumn{7}{c}{Night-Rainy}  \\
                             Method  & Bus & Bike & Car & Motor & Person & Rider & Truck 
                               & Bus & Bike & Car & Motor & Person & Rider & Truck 
                               & Bus & Bike & Car & Motor & Person & Rider & Truck 
                               & Bus & Bike & Car & Motor & Person & Rider & Truck \\
        \midrule
        S-DGOD    & 4.8 & -1.7 & -0.9 & 3.5 & -0.7 & 2.7 & 2.6 & 5.9 & 3.1 & -5.9 & 5.9 & -2.7 & 4.5 & 4.8 & 4.8 & -1.7 & -0.9 & 3.5 & -0.7 & 2.7 & 2.6 & 5.9 & 3.1 & -5.9 & 5.9 & -2.7 & 4.5 & 4.8 \\
        SRCD     & 5.7 & 3.4 & 2.7 & 5.1 & 3.5 & 4.6 & 4.5 
                 & 5.4 & 1.9 & 2.8 & 4.7 & 3.3 & 2.9 & 2.1& 5.7 & 3.4 & 2.7 & 5.1 & 3.5 & 4.6 & 4.5 
                 & 5.4 & 1.9 & 2.8 & 4.7 & 3.3 & 2.9 & 2.1  \\
        \midrule
        GoDiff      & 1.9 & 0.1 & \textbf{5.9} & 0.4 & 1.5 & 1.8 & 2.8 & 1.1 
                    & \textbf{8.3} & 2.0 & 3.9 & \textbf{5.1} & \textbf{3.6} & \textbf{4.3}& \textbf{9.8} & \textbf{6.1} & \textbf{6.6 }& 2.7 & \textbf{6.7} & 3.6 & \textbf{7.8} & \textbf{6.9} 
                    & 1.9 & \textbf{6.1} & -0.3 & 0.2 & -0.1 & 4.3 \\
        \shline
    \end{tabular}
    }
\end{table*}

We conduct ablation studies using the Faster R-CNN detector to evaluate our GoDiff method, comparing various combinations of source and generated domains. Table~\ref{tab:domain} presents the mAP results for these domain configurations, with generated domains listed in rows and target test domains in columns. The baseline row (``/") represents training solely on the source domain (Daytime-Sunny).

Our findings indicate that each generated domain substantially improved detector performance in its corresponding target domain. For instance, the generated Night-Sunny domain increased mAP on the Night-Sunny test set from 31.8\% to 32.6\%, and the generated Daytime-Foggy domain enhance performance on the Daytime-Foggy test set from 32.0\% to 35.0\%. This consistent improvement across all generated domains confirms the effectiveness of our approach.

Moreover, models in some generated domains demonstrate strong generalization to related conditions. For instance, the model in the Dusk-Rainy domain not only improved performance under Dusk-Rainy conditions (from 26.0\% to 32.6\%) but also enhanced results in Night-Rainy scenarios (from 12.1\% to 13.9\%). This suggests that our method effectively captures common features across domains, thereby increasing the detector's overall robustness.

Training with all generated domains combined (``All" row) achieves the best overall performance, attaining the highest mAP in three out of five test domains and the second-highest in the remaining two.

\subsubsection{Analysis of Domain Difference}

\begin{figure}[!t]
\centering 
\includegraphics[width=\linewidth]{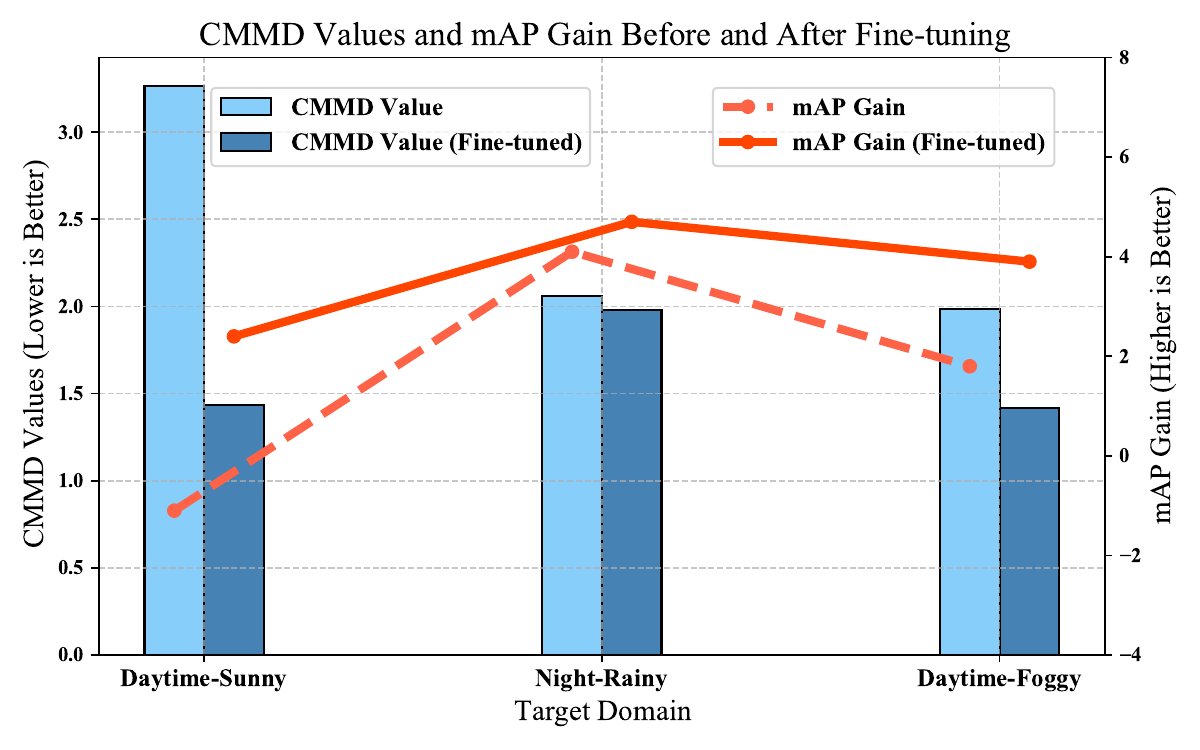}
\caption{The Impact of source domain fine-tuning on generation domain authenticity (CMMD) and target domain performance (mAP).} 
\label{fig:cmmd}
\end{figure}

To evaluate the effectiveness of our GoDiff method, we conduct a detailed analysis focusing on the impact of fine-tuning the diffusion model with source domain data. The analysis centers on two key metrics: the CMMD between generated and real domains, and the resulting improvement in mean mAP for object detection. The findings are illustrated in Figure \ref{fig:cmmd}, which compares these metrics across three domains—Daytime-Sunny, Night-Rainy, and Daytime-Foggy—both before and after fine-tuning.

CMMD values indicate the dissimilarity between the generated pseudo-target domain data and the real target domain data, with lower values reflecting higher similarity. Our results show that fine-tuning the diffusion model with source domain data consistently reduces CMMD values across all three domains. This reduction is especially notable in the Daytime-Sunny scenario, where the CMMD value decreases from 3.3 to 1.4, signifying a substantial improvement in the realism of the generated data. Additionally, we observe a strong correlation between the reduction in CMMD values and the increase in mAP. In all three target domains, the fine-tuned model consistently achieves higher mAP gains compared to the non-fine-tuned version.

Our analysis reveals a clear trend: lower CMMD values, which indicate greater similarity between generated and real data, are associated with larger mAP gains. This finding emphasizes the importance of generating high-quality, domain-specific data to enhance object detection performance across diverse environments.
These results underscore the effectiveness of our approach in improving domain generalization for object detection tasks. By fine-tuning the diffusion model with source domain data, we not only reduce the domain gap between generated and real data but also significantly boost detection performance in challenging scenarios.

Looking forward, these findings suggest a promising direction for future research. Further enhancing the realism of generated domain data could potentially lead to even greater improvements in generalization performance, highlighting the value of refining our data generation techniques.

\subsection{Visualization of Generated Domains}

\begin{figure*}[!t]
\centering 
\includegraphics[width=0.8\linewidth]{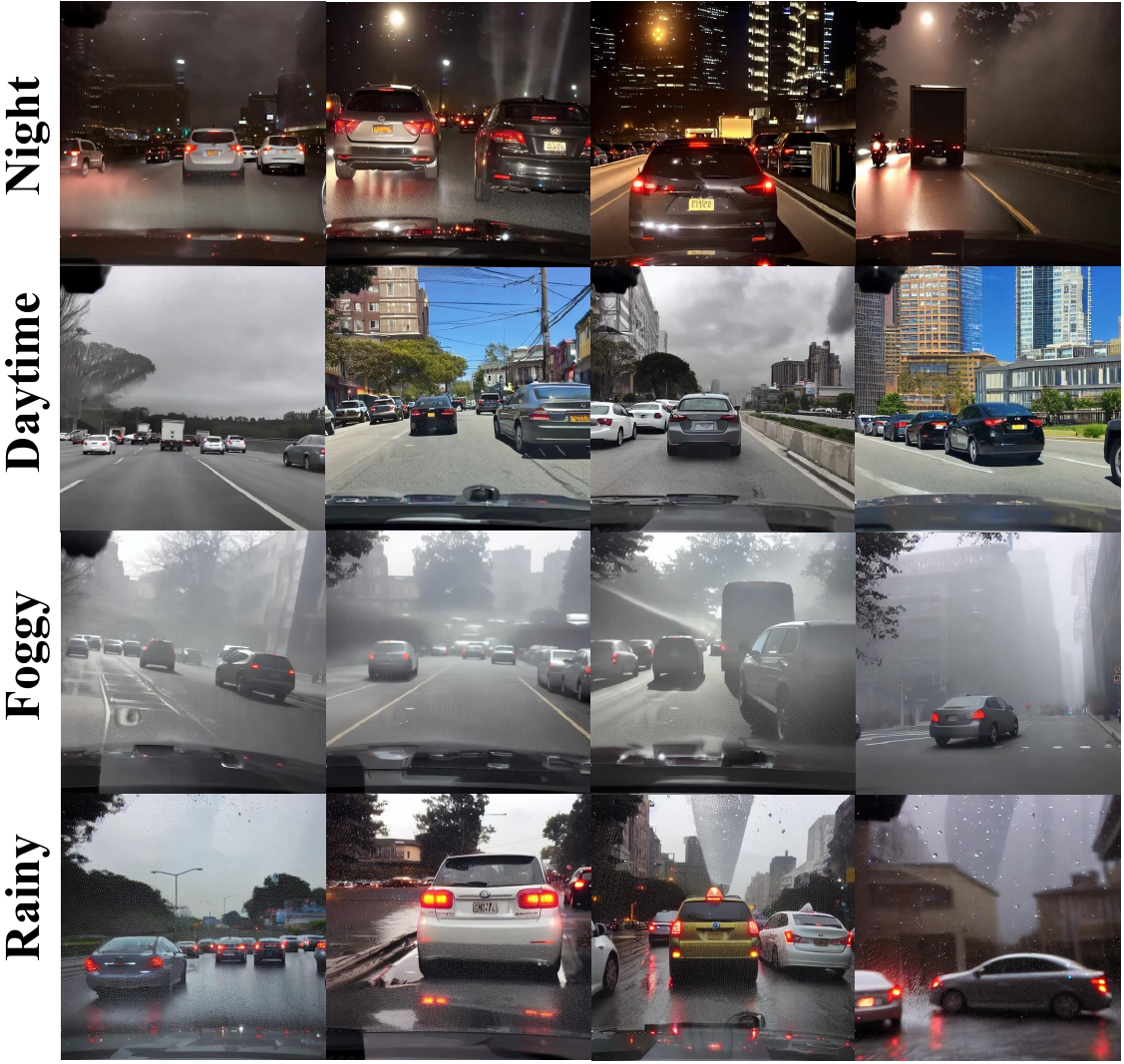}
\caption{Sample images generated by our method across four weather conditions.} 
\label{fig:generateimage}
\end{figure*}

Figure \ref{fig:generateimage} presents sample images generated by our method across four environmental conditions: night, daytime, foggy, and rainy. These visualizations demonstrate the effectiveness of our PTDG module in creating diverse and realistic data.
By applying various visual styles to source images while maintaining their core structure and object relationships, our approach significantly expands the training dataset. This increased diversity is key to improving our object detection model's performance across a range of environmental conditions.

\subsection{Qualitative Results}
\begin{figure*}[!t]
\centering 
\includegraphics[width=0.8\linewidth]{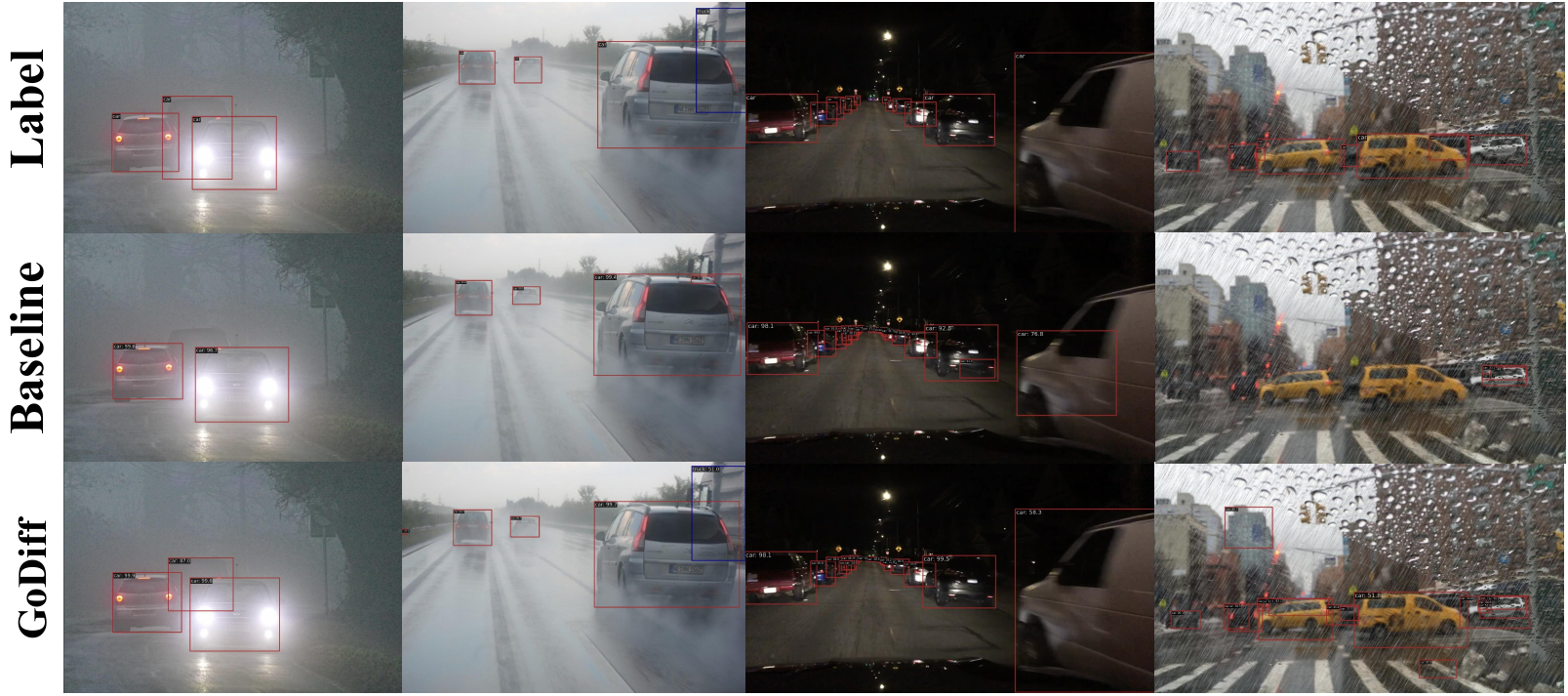}
\caption{Comparison of object detection performance in challenging weather conditions. Rows show ground truth labels (top), baseline detector results (middle), and our proposed method (bottom) across foggy, rainy, nighttime, and urban rainy scenes.} 
\label{fig:sdg}
\end{figure*}

Figure \ref{fig:sdg} showcases our method's improved generalization performance under challenging weather and lighting conditions. The figure compares ground truth annotations, baseline detector results, and outcomes from our proposed generalization method across four scenarios: heavy fog, rain, nighttime, and urban rainy scenes. Our method demonstrates notable improvements over the baseline detector in all tested conditions. 



